\pdfoutput=1

\documentclass[11pt,table]{article}

\usepackage{ACL2023}

\usepackage{times}
\usepackage{latexsym}

\usepackage[T1]{fontenc}

\usepackage[utf8]{inputenc}

\usepackage{microtype}

\usepackage{inconsolata}

\usepackage{graphicx}
\usepackage{xcolor}

\usepackage{subcaption}

\usepackage{booktabs}
\usepackage{multirow}
\usepackage{colortbl}

\usepackage{amsmath,amssymb,amsthm}
\usepackage{bm}

\def\eqref#1{equation~\ref{#1}}

\def\1{\bm{1}}

\def\vtheta{{\bm{\theta}}}
\def\va{{\bm{a}}}
\def\vb{{\bm{b}}}
\def\vc{{\bm{c}}}

\def\vn{{\bm{n}}}
\def\vo{{\bm{o}}}

\def\vr{{\bm{r}}}
\def\vs{{\bm{s}}}

\def\vx{{\bm{x}}}

\def\mA{{\bm{A}}}

\def\mR{{\bm{R}}}

\DeclareMathAlphabet{\mathsfit}{\encodingdefault}{\sfdefault}{m}{sl}
\SetMathAlphabet{\mathsfit}{bold}{\encodingdefault}{\sfdefault}{bx}{n}

\def\sA{{\mathbb{A}}}

\def\sN{{\mathbb{N}}}

\def\sS{{\mathbb{S}}}

\def\sY{{\mathbb{Y}}}

\usepackage[ruled,vlined,linesnumbered]{algorithm2e}
\usepackage{ragged2e}

\newcommand{\vect}[1]{\ensuremath{\bm{#1}}} %
\newcommand{\lang}[1]{\ensuremath{\textit{#1}}}

\newif\ifcomments
\commentstrue

\ifcomments\newcommand{\comments}[1]{#1}\else\newcommand{\comments}[1]{}\fi

\title{Does CLIP Bind Concepts? \\ Probing Compositionality in Large Image Models}

\author{
Martha Lewis$^{1}$\thanks{\; Equal contribution}\ \;\;\; 
Nihal V. Nayak$^{2 *}$\;\;\; 
Peilin Yu$^{2}$ \;\;\; 
\textbf{Qinan Yu}$^{2}$\;\;\; 
\textbf{Jack Merullo}$^{2}$\\
\textbf{Stephen H. Bach}$^{2}$\;\;\; 
\textbf{Ellie Pavlick}$^{2}$\\
$^1$ School of Engineering Mathematics and Technology, University of Bristol\\
$^2$ Department of Computer Science, Brown University\\
\texttt{martha.lewis@bristol.ac.uk}, \texttt{nihal\_vivekanand\_nayak@brown.edu}\\ \{\texttt{peilin\_yu}, \texttt{qinan\_yu}, \texttt{jack\_merullo}, \texttt{stephen\_bach}, \texttt{ellie\_pavlick}\}\texttt{@brown.edu}
}

\begin{document}
\maketitle
\begin{abstract}
Large-scale neural network models combining text and images have made incredible progress in recent years. However, it remains an open question to what extent such models encode compositional representations of the concepts over which they operate, such as correctly identifying \lang{red cube} by reasoning over the constituents \lang{red} and \lang{cube}. In this work, we focus on the ability of a large pretrained vision and language model (CLIP) to encode compositional concepts and to bind variables in a structure-sensitive way (e.g., differentiating \lang{cube behind sphere} from \lang{sphere behind cube}). To inspect the performance of CLIP, we compare several architectures from research on compositional distributional semantics models (CDSMs), a line of research that attempts to implement traditional compositional linguistic structures within embedding spaces. We benchmark them on three synthetic datasets -- single-object, two-object, and relational -- designed to test concept binding. We find that CLIP can compose concepts in a single-object setting, but in situations where concept binding is needed, performance drops dramatically. At the same time, CDSMs also perform poorly, with best performance at chance level.
\end{abstract}

\section{Introduction}

Good semantic representations are generally assumed to require, at a minimum, \textit{compositionality} and \textit{groundedness}. That is, meanings of sentences should be functions of the words they contain and the syntax via which those words are combined \cite{partee1995lexical} (\textit{compositionality}), and such meanings should be at least in part responsible for reference to the real world, e.g., via truth conditions (\textit{groundedness}). The current state-of-the-art of semantic representation consists of vectors extracted from very large neural networks trained either on text alone \cite{devlin2019,brown:neurips20,touvron2023llama} or a mix of text and images \cite{radford2021a,openai:arxiv23}. It remains a wide-open question whether such models constitute good semantic representations \cite{pavlick2022semantic}, with empirical evidence and in-principle arguments simultaneously supporting claims that models are and are not compositional \cite{compositionalintelligence:23}, and that they are and are not grounded \citep{piantadosi:arxiv22,bender:acl22,mollo2023vector}. 

In this paper, we focus on vision-and-language models\footnote{There is significant debate about whether text-only language models can be considered ``grounded''. It is often assumed that models trained on multimodal data will circumvent this debate, but this should not be taken for granted. Our findings add to work which shows that VLMs don't necessarily learn a grounded semantics of the type traditionally sought in linguistics; further work and debate is necessary to make normative claims about the representations that VLMs learn.} 
(specifically CLIP and fine-tuned variants of CLIP), and seek to answer, in a controlled setting, whether such models meet basic tests of grounded compositionality. Specifically, we consider three basic types of linguistic compositions: combining a single adjective and noun (\textit{red cube}), combining two adjectives with respective nouns (\textit{red cube and blue sphere}), and relating two nouns (\textit{cube behind sphere}). All three of these settings require some degree of compositionality and groundedness, with the latter two exemplifying a more abstract type of compositionality (pervasive in language) which depends not only on recognizing a conjunction of constituents but an ability to bind meaning representations to abstract syntactic roles. 
Recently, there has been a significant interest in the community to benchmark the compositional capabilities of CLIP and other VLMs \citep{ma:arxiv22, yuksekgonul:iclr23,thrush:cvpr22}. 
However, \citet{hsieh2023} shows that these datasets are `hackable' as the incorrect labels may not be meaningful and do not require the image to predict the correct label.
For example, an image of \lang{a horse eating the grass} can have the distractor \lang{the grass eating a horse}. 
In contrast, we are less prone to such ``hackable'' artifacts as we include meaningful distractors that require both the image and the labels for the final prediction.  
We therefore provide a controlled setting for benchmarking compositionality in CLIP.

We situate our work within the tradition of research on \textit{compositional distributional semantics models} (CDSMs) \cite{erk2008,mitchell2010,baroni2010, coecke2010,boleda2020}, which seek to bridge the gap between distributional models and formal semantics by building architectures which operate over vectors yet still obey traditional theories of linguistic composition.

Formal semantics approaches such as \citet{montague1973} describe how the meaning of a sentence can be built from its component parts. This approach to meaning representation accounts for how a wide variety of expressions can be produced by speakers, and how we can understand sentences that we have never heard before by composing their component parts. Phenomena such as inference are also easily accounted for -- although there are still difficulties~\cite{partee1995lexical}.

Distributional semantics approaches represent word meanings according to their distribution in large text corpora. These have been extremely successful in encoding lexical meaning \cite{landauer1997,mikolov2013a}, as well as in a variety of applications \citep{turney2010}.

CDSMs unify these approaches by representing the symbolic, compositional structure of formal semantic models within vector spaces. This allows for the principled compositional approaches seen in formal semantics to be applied within the distributional setting, using lexical meaning representations from the latter  arena.

CDSMs are intrinsically compositional, and because of this, they have the potential to model concept binding effectively. CDSMs also have the capacity to capture a range of linguistic and cognitive phenomena \cite{smolensky2012symbolic}, and lend themselves to modeling the truth value as well as the meaning of sentences \cite{emerson-copestake-2016-functional}, or accounting for polysemy \cite{boleda2020}. Because of their formal background, they are also potentially more interpretable than current large language models.

We adapt several CDSMs to the grounded language setting, and compare the performance of CLIP's text encoder (tuned in various settings) to the performance of these explicitly compositional models. Overall, we see that on single adjective-noun compositions (\textit{red cube}), CLIP performs better than any of the more explicitly compositional CDSMs. In the other settings, which rely on the ability to bind variables, we see that using CDSMs for the text encoder sometimes improves performance, but not always, and that, across all models, performance is essentially at chance in the best case. These results suggest that CLIP's representation of the visual world is poorly suited for compositional semantics, and suggest that future work on improving these representations is a necessary next step in advancing work on grounded compositional distributional semantics. 

In summary, we make the following contributions:
\begin{itemize}
    \item We provide a controlled analysis of the ability of CLIP and fine-tuned variants to perform compositional visual reasoning tasks.
    \item We adapt a variety of traditional compositional distributional semantics (CDS) architectures to the grounded language setting.
    \item We show that all our models perform poorly on generalization settings that require abstract variable binding, suggesting major limitations in the way CLIP represents the visual world. 
\end{itemize}

 \section{Models}\label{sec:models}
In this work, we are interested in comparing contemporary ``end to end'' methods for training neural networks with explicitly compositional models of the type developed in compositional distributional semantics \cite{erk2008, mitchell2010, baroni2010,coecke2010, boleda2020} (henceforth CDSMs for ``compositional distributional semantics models''). Below, we describe the models we compare, including baselines, explicitly compositional models, and contemporary vision-and-language models. 

\subsection{Setup}
\label{sec:setup}
We describe a unified setup that we use to represent compositions in CLIP-based models as well as in CDSMs.
For each compositional task, we are given a dataset $\sS=\{(x_{1},y_{1}),\dots,(x_{N},y_{N})\}$ where $x$ is the image and $y\in\sY$ is a \textit{phrase} which correctly describes the image where $\sY$ is the set of all phrases.
We use CLIP~\citep{radford2021a} to get image embeddings for all input images. 
Embeddings for the phrases are generated either using the text encoder in CLIP (possibly fine-tuned) or using CDSMs.

We train different CLIP variants and CDSMs in order to encode each of the phrases. 
We deal with two types of phrases, namely, adjective-noun and subject-relation-object phrases. 
Let $\sA=\{a_{1},\dots,a_{n}\}$ be the adjectives and $\sN=\{n_{1},\dots,n_{m}\}$ be the nouns in an adjective-noun phrase. 
The models produce the adjective-noun phrase embedding $\mathcal{T}(a,n)$ in the joint semantic space where $a\in\sA$ and $n\in\sN$. 
Letting $\mathrm{R}=\{\mathcal{R}_{1},\dots,\mathcal{R}_{n}\}$ be the relations, the model generates the relational phrase embedding $\mathcal{T}(s, \mathcal{R}, o)$ where the subject is $s\in\sN$, the relation is $\mathcal{R}\in\mathrm{R}$, and the object is $o\in\sN$.
All models, with the exception of frozen CLIP, are trained to update phrase embeddings based on the training data. For the compositional models, the word embeddings that are composed to form the phrase embedding are updated.
For more details, see Section \ref{sec:experiments}.

\subsection{CLIP and Variants}\label{sec:clip_variants}
We examine the performance of CLIP \citep{radford2021a}, fine-tuned CLIP, and a compositional variant~\citep{nayak2022} on the tasks.

\paragraph{CLIP}
CLIP \citep{radford2021a} is a pretrained vision-and-language model trained with a contrastive loss objective on 400 million image-text pairs. 
The architecture includes two key components: an image encoder and a text encoder that produce vector representations for images and texts in the joint semantic space.
The text encoder accepts prompts in natural language to produce zero-shot classifiers.
We get the final prediction by taking the cosine similarity between the image and the text vectors and choosing the text with the highest similarity score. 
This ability enables us to test CLIP out-of-the-box on compositional tasks.
We set the following prompt templates for the adjective-noun and subject-relation-object setting: 
 \begin{align*}
 	\mathcal{T}(a, n)&= \phi(\text{\texttt{a photo of adj noun}}) \\
	\mathcal{T}(s, \mathcal{R}, o) &= \phi(\text{\texttt{a photo of sub rel obj}})
\end{align*}
where $\phi$ is the CLIP pretrained text encoder, \texttt{adj noun} is replaced with the adjective and noun pairs, and \texttt{sub rel obj} is replaced with nouns and relations from the dataset. 
We consider frozen CLIP and a fine-tuned variant CLIP-FT (Section \ref{sec:experiments}). 

\paragraph{Compositional Soft Prompting}
CSP or compositional soft prompting \citep{nayak2022} is a parameter-efficient learning technique designed to improve the compositionality of large-scale pretrained models like CLIP. 
They focus on real-world adjective-noun datasets which contain images of a single object associated with an adjective.
They fine-tune embeddings of tokens corresponding to adjective and object concepts on a set of seen classes while keeping other parameters of the text and the image encoders frozen.
During inference, they recompose adjective and object tokens in new concatenations for zero-shot inference.
In this work, we systematically evaluate CSP on different types of compositional tasks (Section \ref{sec:experiments}).
We set the following prompt templates for the adjective-noun and subject-relation-object setting: 
 \begin{align*}
 	&\mathcal{T}(a, n)= \phi(\text{\small\texttt{a photo of [adj] [noun]}}) \\
	&\mathcal{T}(s, \mathcal{R}, o) = \phi(\small\text{\texttt{a photo of [sub] [rel] [obj]}})
\end{align*}
where $\phi$ is the pretrained text encoder in CLIP, \texttt{[adj] [noun]} are the fine-tuned token embeddings for adjectives and nouns and \texttt{[sub] [rel] [obj]} are the fine-tuned token embeddings for nouns and relations in the dataset.

\subsection{Compositional Distributional Semantics Models (CDSMs)}\label{sec:cdsm}
We consider a number of compositional distributional semantics models, which have been proposed in past work but have not been applied to a grounded language setting. 
Each of these models trains embeddings (vectors, matrices, or tensors) for each word in the class, and then composes them together to produce a compositional phrase embedding. 
All models are trained to learn the phrase embeddings by aligning them with the frozen image embeddings from CLIP.

\paragraph{Syntax Insensitive Models (Add, Mult, Conv)}
We consider three simple compositional models that are insensitive to order. 
The first two are Add, consisting of combining word vectors by addition, and Mult, where word vectors are combined by pointwise multiplication \citep{mitchell2010,grefenstette2011a}.  
Lastly, we use circular convolution (Conv) \citep{plate1995}. 
For $\va$, $\vb$, $\vc \in \mathbb{R}^n$,  $\vc = \mathrm{Conv}(\va, \vb) = \va \circledast \vb$ means that $ \vc_i= \sum_{j=0}^{n-1} \va_j\vb_{i-j}$ where $i-j$ is interpreted as modulo $n$.
 
\begin{figure*}[t]
    \centering
    \begin{subfigure}[t]{0.3\textwidth}
    \centering
        \includegraphics[width=1\linewidth]{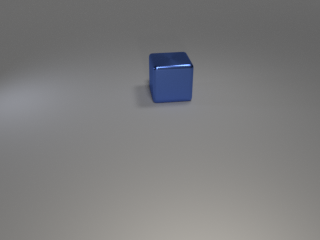}
        \caption{Single-object dataset. Example true label and distractors are:
        \{\textcolor{teal}{blue cube}, \textcolor{red}{yellow sphere}, \textcolor{red}{gray cube}, \textcolor{red}{purple cylinder}, \textcolor{red}{cyan cylinder}\}}
        \label{fig:single}
    \end{subfigure}
    \hfill
    \begin{subfigure}[t]{0.3\textwidth}
    \centering
        \includegraphics[width=1\linewidth]{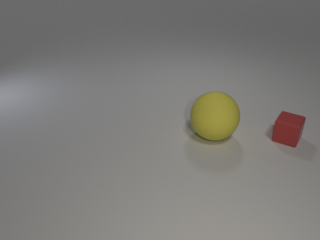}
        \caption{Two-object dataset. Example true label and distractors are:
        \{\textcolor{teal}{yellow sphere}, \textcolor{red}{yellow cube}, \textcolor{red}{red sphere}, \textcolor{red}{blue cube}, \textcolor{red}{purple cylinder}\}. \lang{yellow cube} and \lang{red sphere} are `hard' distractors.}
        \label{fig:two}
    \end{subfigure}
    \hfill
    \begin{subfigure}[t]{0.3\textwidth}
        \centering
        \includegraphics[width=1\linewidth]{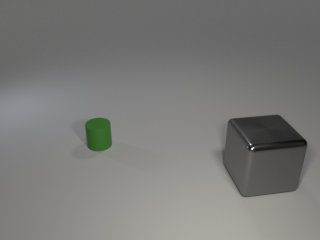}
        \caption{Relational dataset. Example true label and distractors are:
        \{\textcolor{teal}{cylinder left of cube}, \textcolor{red}{cube left of cylinder}, \textcolor{red}{cylinder right of cube}, \textcolor{red}{sphere left of cube}, \textcolor{red}{cylinder left of sphere}\}.}
        \label{fig:rel}
    \end{subfigure}
    \caption{Example images and label sets from each dataset. 
    The texts in \textcolor{teal}{Green} are the true classes and \textcolor{red}{Red} are the distractors. 
    Unlike the two-object and relational datasets, the single-object dataset does not require concept binding. 
    }\label{fig:images}
\end{figure*}

\begin{table*}[t!]
    \centering
    \begin{tabular}{lccccccccc}\toprule
    && \multicolumn{2}{c}{{Train}} && \multicolumn{2}{c}{{Validation}} && \multicolumn{2}{c}{{Generalization}} \\\cmidrule{3-4}\cmidrule{6-7}\cmidrule{9-10}
    {Dataset} && {\# Examples} & {\# Classes} && {\# Examples} & {\# Classes} && {\# Examples} & {\# Classes}\\\cmidrule{1-1}\cmidrule{3-4}\cmidrule{6-7}\cmidrule{9-10}
    Single-object && 5598 & 14 && 799 & 2 && 3195 & 8 \\
    Two-object && 20000 & 14 && 20000 & 2 && 20000 & 8 \\
    Relational && 40000 & 20 && 20000 & 2 && 20000 & 2 \\
    \bottomrule
    \end{tabular}
    \caption{
    Summary of the statistics of the datasets in the concept binding benchmark. 
    }
    \label{tab:datasets}
\end{table*}

 \paragraph{Type-logical model (TL)}
Type-logical approaches to distributional semantics map grammatical structure into vector space semantics~\citep{baroni2010,coecke2010}. 
Concretely, we represent the nouns as vectors, adjectives as matrices, and the composition of an adjective and a noun is given by matrix-vector multiplication. 
Following \citet{kartsaklis2012}, we represent transitive verb or relation as a matrix, and the composition of the noun-relation-noun is given by matrix-vector multiplication followed by pointwise vector multiplication, i.e.:
\begin{align*}
\mathcal{T}(a, n) &= \mA\cdot\vn, \quad \mathcal{T}(s, \mathcal{R}, o) =  \vs \odot (\mR \cdot \vo)
\end{align*}
where $\vn$, $\vs$, and $\va$ are learnable embeddings, $\mA$ and $\mR$ are learnable weight matrices, $\cdot$ is matrix-vector multiplication and $\odot$ is pointwise multiplication . 

\paragraph{Role-filler model (RF)}
Introduced in \citet{smolensky1990}, role-filler-based representations provide a means of representing structure using vectors. 
A symbolic structure can be represented as a collection of role-filler bindings, instantiated within a vector space.
Consider \lang{red cube} which is rendered as $\mathbf{red} \circledast \mathbf{adj.} + \mathbf{cube} \circledast \mathbf{noun}$ where $\mathbf{adj.}$ and $\mathbf{noun}$ are role vectors, $\mathbf{red}$ and $\mathbf{cube}$ are filler vectors, and circular convolution $\circledast$ is a binding operator \cite{plate1995}.
Formally, we learn an embedding for each filler, of type noun, adjective, or relation, and another set of embeddings for each role: 
\begin{align*}
\nonumber \mathcal{T}(a, n) &= \va\circledast \vr_a + \vn\circledast \vr_n\\
\nonumber \mathcal{T}(s, \mathcal{R}, o) &=  \vs \circledast \vr_s + \vect{R} \circledast \vr_R + \vo \circledast \vr_o
\end{align*}
where all of $\va$, $\vn$, $\vs$, $\vect{R}$, $\vo$, $\vr_a$, $\vr_n$, $\vr_s$, $\vr_R$, and $\vr_o$ are learnable embeddings and $\circledast$ is the circular convolution operation.

\section{Concept Binding Benchmark}\label{sec:datasets}
We introduce the concept binding benchmark to evaluate the compositional generalization capabilities of VLMs. 
In this benchmark, we introduce three datasets: single-object, two-object, and relational (see Figure \ref{fig:images}). 
Following \citet{johnson2017}, we use \citet{blender:2018} to generate synthetic datasets with objects of simple shapes and colors.
Each dataset contains train, validation, and generalization sets with no overlap in the true class labels. 
Class labels are of the form  \lang{adjective-noun} or \lang{subject-relation-object}. 
All individual nouns, adjectives, and relations are included in the training sets such that we can train models on the training set and test for compositional generalization on held-out classes in the validation and generalization set. 
Unlike prior work that introduces datasets with a focus on concept binding \citep{yuksekgonul:iclr23,ma:arxiv22,thrush:cvpr22}, our synthetically generated datasets contain both semantically meaningful and hard labels and provide a controlled setting to evaluate the compositional capabilities of VLMs. 
Table \ref{tab:datasets} shows the statistics of the datasets.

\paragraph{Single-object dataset}
\label{sec:single_obj}
The dataset consists of images of exactly one object of a given shape and color (see Figure \ref{fig:single}). 
We consider the following shapes and colors: cubes, spheres, and cylinders and blue, gray, yellow, brown, green, purple, red, and cyan with a total of $24$ possible combinations. 
The validation set includes brown cube and green cylinder and the generalization set includes green cube, purple cube, red cube, cyan cube, blue cylinder, gray cylinder, yellow cylinder, and brown cylinder. 
The remainder of the combinations are included in the training set. 
The correct label for the image is an adjective-noun label. Four distractors are sampled from the other possible adjective-noun combinations. 

\paragraph{Two-object dataset}\label{sec:attr_two_obj_data}
The dataset contains images with two objects of different shapes each associated with a different color (see Figure \ref{fig:two}).
Following the single object experiments, we use the same shape-color combinations in the train, validation, and generalization split. 
A correct label for a given image is again an adjective-noun label.
However, we manually choose ``harder'' distractors by switching the adjective and object compositions. 
For example, in Figure \ref{fig:two} we have two classes \lang{red cube} and \lang{yellow sphere}.
When \lang{red cube} is the positive label, we set two of the four distractors to be \lang{red sphere} and \lang{yellow cube}. 
The other two distractors are randomly sampled from the pool of negative labels, say \lang{blue sphere} and \lang{red cylinder}. 
We follow the same procedure when \lang{yellow sphere} is the positive example. 

\paragraph{Relational dataset}\label{sec:rel_two_obj_data}
This dataset contains images with two objects. A correct label for an image is given by a phrase of the form \lang{subject relation object}.
We consider the following objects and relations: cube, sphere, and cylinder and left, right, front, and behind. 
This means there are 24 possible combinations of spatial relations of the form $a\mathcal{R}b$ where $\{a, b\}$ are objects and  $a\neq b$ and $\mathcal{R}$ is the relation. 
For each image, the distractor labels are constructed as $\{b\mathcal{R}a, a\mathcal{S}b, a\mathcal{R}c, c\mathcal{R}b\}$ where $c \notin \{a, b\}$ is an object type other than $a$ or $b$ and $\mathcal{S}$ is the relation opposite to $\mathcal{R}$. 
The validation set includes images of cubes in front of spheres (equivalently, spheres behind cubes), and the generalization set includes images of cylinders in front of cubes (equivalently, cubes behind cylinders). 
All the other 20 image types are seen in the training set, and note that shapes can appear on either side of the image. 
Figure \ref{fig:rel} shows an example from the training set with a \lang{cylinder behind cube}. 

 \section{Experiments and Results}\label{sec:experiments}
To understand the compositional capabilities of CLIP, we benchmark CLIP and the compositional models from Section \ref{sec:models} on the three datasets described in Section \ref{sec:datasets}.
Detailed training setup and parameters are given in Appendix \ref{sec:training_details}.
We have released code and datasets for all experiments.\footnote{\url{https://github.com/marthaflinderslewis/clip-binding}}

\subsection{Single Adjective-Noun Composition}
We test the ability of our models to correctly classify the composition of objects with properties  (e.g., ``red cube'') in the single-object dataset.

\begin{table}[t!]
    \centering
    \resizebox{1.0\linewidth}{!}{
    \begin{tabular}{lrrr}
    \toprule
    Model & Train & Val & Gen \\\midrule
   CLIP & 94.23 & 97.75 & 92.39\\
    \midrule
    CLIP-FT&  98.98 $_{1.02}$&  89.06 $_{5.84}$&  78.54 $_{4.41}$\\
    CSP & 94.98 $_{0.45}$&  84.58 $_{0.16}$&  88.74 $_{0.34}$\\
   \midrule
{Add}&  99.77 $_{0.03}$&  44.98 $_{1.32}$&  85.16 $_{0.96}$\\
{Mult}&  43.27 $_{13.9}$&  4.48 $_{4.08}$&  5.38 $_{2.66}$\\
{Conv}&  41.10 $_{14.3}$&  7.33 $_{2.90}$&  4.11 $_{1.53}$\\
{TL}&  99.98 $_{0.02}$&  1.08 $_{0.44}$&  0.92 $_{0.24}$\\
{RF}&  98.87 $_{0.11}$&  59.52 $_{6.12}$&  80.64 $_{1.36}$\\
    \bottomrule
    \end{tabular}
    }
    \caption{
    Results for all models on single adjective-noun composition, training epoch chosen by performance on validation set.
    We report the average accuracy for all the methods on 5 random seeds and the standard error.}
    \label{tab:single_acc}
\end{table}

\begin{table}[t!]
    \centering
    \resizebox{1.0\linewidth}{!}{
    \begin{tabular}{lrrr}
    \toprule
Model & Adj & Noun & Both\\\midrule
CLIP &  83.47               & 14.87              &1.65\\\midrule
CLIP-FT &  0.12 $_{ 0.12}$ &  92.95 $_{ 4.09}$ &  6.94 $_{ 3.98}$\\
CSP &  85.19 $_{ 0.72}$ &  12.57 $_{ 0.72}$ &  2.24 $_{ 0.05}$\\ \midrule
Add &  94.85 $_{ 0.51}$ &  1.13 $_{ 0.22}$ &  4.02 $_{ 0.43}$\\
Mult &  33.47 $_{ 3.17}$ &  14.70 $_{ 2.62}$ &  51.84 $_{ 5.75}$\\
Conv &  29.59 $_{ 3.19}$ &  13.12 $_{ 1.84}$ &  57.29 $_{ 4.25}$\\
 TL &  39.18 $_{ 0.72}$ &  21.64 $_{ 0.27}$ &  39.17 $_{ 0.50}$\\
RF &  64.01 $_{ 2.70}$ &  10.99 $_{ 1.08}$ &  24.99 $_{ 2.50}$\\
\midrule
    \end{tabular}}
    \caption{
    Percentages assigned to each type of error for the single-object color task, generalization split. Here, Adj means the model predicted the adjective incorrectly but the noun correct; Noun means the opposite error; and Both means the model predicted neither the adjective nor the noun correctly.  We report the average error proportions for all the methods on 5 random seeds and the standard error.}
    \label{tab:single-obj-err}
\end{table}

\paragraph{Results}

In Table \ref{tab:single_acc}, we see that frozen CLIP outperforms all the models.
CLIP achieves 97.75\% on the validation set and 92.39\% on the generalization set. 
After fine-tuning, CLIP's performance drops to 89.06\% on the validation set and 78.54\% on the generalization set. 
We observe a similar trend in CSP, i.e., the performance on the validation set reduces to $84.58\%$  but achieves slightly better performance on the generalization set with 88.74\%.
We suspect this drop is because the model overfits to the true compositions in the training set.\footnote{
Calibrating predictions on the validation set is a common practice in zero-shot learning to reduce bias towards seen classes. 
We find calibration improves CSP from 88.74\% to 96.31\% on the single-object setting. 
This shows fine-tuned variants of CLIP can generalize better than frozen CLIP. 
However, calibration in the two-object setting does not improve generalization accuracy suggesting this setting is harder as it requires \lang{binding} adjectives to objects. 
Details in Appendix \ref{sec:calibrated_stacking}.
}
Out of the CDSMs, Add and RF both perform well on training and generalization sets, achieving 80.64\% and 85.16\% on the generalization set respectively. 
We see that Conv, Mult, and TL are unable to generalize to the validation and the generalization sets. 
These three models can achieve high performance (high 90s) on the training set after several epochs but at the expense of performance on the validation set (not included in Table \ref{tab:single_acc} as we report accuracy based on best performance on the validation set). 

A breakdown of errors on the generalization set is reported in Table \ref{tab:single-obj-err}. 
We see that CSP, Add, and RF have similar types of errors, i.e., these models often predict the incorrect adjective but predict the correct noun. 
CLIP-FT, however, predicts the adjective (color) correctly but gets the noun wrong.

\subsection{Two-Object Adjective-Noun Binding}\label{sec:two_object}
In this task, we test whether CLIP can \textit{bind} concepts together.
Given two objects, can CLIP bind adjectives to correct objects as opposed to merely representing the image as a ``bag of concepts''?
For example, in Figure \ref{fig:two}, can CLIP predict that the image contains a \lang{red cube} rather than a \lang{yellow cube}?

\begin{table}[t!]
    \centering
    \resizebox{1.0\linewidth}{!}{
    \begin{tabular}{lrrr}
    \toprule
    Model & Train & Val & Gen \\
    \midrule
   CLIP & 27.02 & 7.17 & 31.40 \\
    \midrule  
CLIP-FT &  86.91 $_{8.15}$&  6.31 $_{3.31}$&  0.25 $_{0.10}$\\
CSP&  37.59 $_{1.54}$&  20.98 $_{0.22}$&  11.15 $_{2.03}$\\
    \midrule
Add&  32.46 $_{0.11}$&  15.38 $_{0.89}$&  21.37 $_{0.60}$\\
Mult&  86.65 $_{8.93}$&  4.66 $_{1.35}$&  0.13 $_{0.03}$\\
Conv&  46.26 $_{0.53}$&  7.11 $_{2.18}$&  0.28 $_{0.14}$\\
TL&  99.41 $_{0.17}$&  21.23 $_{4.08}$&  0.08 $_{0.07}$\\
RF&  25.23 $_{1.08}$&  25.13 $_{3.99}$&  20.36 $_{1.36}$\\
    \bottomrule
    \end{tabular}
    }
    \caption{
    Results for all models on adjective-noun binding task, training epoch chosen by performance on validation set. We report the average accuracy for all the methods on 5 random seeds and the standard error.
    }
    \label{tab:two_acc}
\end{table}

\begin{table}[t]
    \centering
    \resizebox{1.0\linewidth}{!}{
    \begin{tabular}{lrrr}
    \toprule
Model&Adj&Noun&Both\\
\midrule
CLIP & 53.08 & 45.40 & 1.51 \\\midrule
CLIP-FT &  47.63 $_{ 0.26}$ &  46.89 $_{ 1.20}$ &  5.48 $_{ 1.01}$\\
CSP &  49.22 $_{ 0.54}$ &  48.25 $_{ 0.72}$ &  2.53 $_{ 0.17}$\\ \midrule
Add &  53.57 $_{ 0.16}$ &  44.32 $_{ 0.25}$ &  2.11 $_{ 0.23}$\\
Mult &  48.51 $_{ 0.03}$ &  46.43 $_{ 1.13}$ &  5.06 $_{ 1.15}$\\
Conv &  44.27 $_{ 0.19}$ &  38.20 $_{ 0.35}$ &  17.53 $_{ 0.43}$\\
TL &  48.76 $_{ 0.03}$ &  47.85 $_{ 0.12}$ &  3.39 $_{ 0.15}$\\
RF &  50.64 $_{ 0.91}$ &  41.32 $_{ 1.26}$ &  8.04 $_{ 1.46}$\\
\bottomrule
    \end{tabular}
    }
    \caption{Percentages assigned to each type of error for the two-object setting.  
    Here, Adj means the model predicted the adjective incorrectly but the noun correct; Noun means the opposite error; and Both means the model predicted neither the adjective nor the noun correctly.
    We report the average error proportions for all the methods on 5 random seeds and the standard error.}
    \label{tab:two_obj_err}
\end{table}

\paragraph{Results}
This task is more challenging for all models (Table \ref{tab:two_acc}). 
Frozen CLIP performs at a level close to chance. 
After fine-tuning, we see that CLIP-FT overfits to the training set, achieving good training accuracy (86.91\%), but falling much lower on validation and generalization (6.31\% and 0.25\% respectively). 
At the epoch with the best accuracy on the validation set, CSP has a lower performance on the training set and slightly higher on the validation and generalization sets compared to CLIP-FT. 
However, as training progresses, we observe that CSP also overfits to the training set (not reported in the table). 
We see that Conv, Mult and TL also exhibit the same pattern of overfitting to the training data, with high training accuracy and low validation and generalization accuracy.
The additive models, Add and RF, underfit the training set and show random accuracy on validation and generalization sets.

Table \ref{tab:two_obj_err} shows that the errors are similar across the models.
For most models, the errors are evenly split between the adjectives and the nouns while only a small proportion of the errors get both incorrect. 
However, we find that Conv incorrectly predicts both the adjective and noun.
For the best performing models, Add and RF, there is a slight bias towards getting the adjective wrong rather than the noun.

 \subsection{Relational Composition}\label{sec:rel}
In this task, we test understanding of spatial relationships between objects, i.e., can our models \textit{bind} objects to positions?
This task requires the models to encode an order or relation between two arguments.
For example, in Figure \ref{fig:rel}, can CLIP differentiate between \lang{cube behind cylinder} and \lang{cylinder behind cube}, even though they have the same words?

\paragraph{Results}
Frozen CLIP performs slightly better than chance on the training set, but worse on the validation and generalization sets, indicating that these may be more difficult (Table \ref{tab:rel_acc}).
After fine-tuning, CLIP-FT improves to around 50\% on the training set, but is completely unable to generalize. This pattern is also seen for CSP and TL.
All the other CDSMs perform slightly above chance. 
This is to be expected for Add, Mult, and Conv because they are commutative.  
Surprisingly, RF is unable to perform better than chance in this setting. 
We suspect that RF has a lower capacity as RF only fine-tunes the role and filler parameters. 
Fine-tuning the image encoder along with the role and filler parameters will increase the complexity of the model and potentially improve the performance on the various splits. 

\begin{table}[t!]
    \centering
    \resizebox{1.0\linewidth}{!}{
    \begin{tabular}{lrrr}
    \toprule
    Model & Train & Val & Gen \\
    \midrule
   CLIP & 26.80 & 14.99 & 0.00 \\
    \midrule
CLIP-FT&  49.59 $_{0.44}$& 0.00 $_{0.00}$& 0.00 $_{0.00}$\\
CSP&  30.40 $_{0.11}$& 0.12 $_{0.01}$& 0.03 $_{0.00}$\\
    \midrule
    Add&  25.41 $_{0.13}$& 26.03 $_{0.07}$& 25.47 $_{0.18}$\\
    Mult&  25.67 $_{0.12}$& 25.95 $_{0.09}$& 25.78 $_{0.09}$\\
    Conv&  24.83 $_{0.06}$& 26.36 $_{0.55}$& 24.95 $_{0.11}$\\
    TL&  67.19 $_{0.26}$& 0.00 $_{0.00}$& 0.00 $_{0.00}$\\
    RF&  25.18 $_{0.28}$& 24.89 $_{0.73}$& 22.78 $_{0.20}$\\
    \bottomrule
    \end{tabular}
    }
    \caption{
    Results for all models on relational composition.
    We report the average accuracy for all the methods on 5 random seeds and the standard error. 
    }
    \label{tab:rel_acc}
\end{table}

\begin{table}[t]
    \centering
    \resizebox{1.0\linewidth}{!}{
    \begin{tabular}{lrrrr}
    \toprule
Model&$b\mathcal{R}a$& $a\mathcal{S}b$& $a\mathcal{R}c$& $c\mathcal{R}b$\\
\midrule
CLIP & 50.00 & 50.00 & 0.00 & 0.00\\\midrule
CLIP-FT &  37.54 $_{ 7.60}$ &  45.97 $_{ 2.41}$ &  12.19 $_{ 7.78}$ &  4.30 $_{ 1.94}$\\
CSP &  49.75 $_{ 0.01}$ &  49.77 $_{ 0.01}$ &  0.40 $_{ 0.01}$ &  0.08 $_{ 0.00}$\\\midrule
Add &  34.21 $_{ 0.08}$ &  65.79 $_{ 0.08}$ &  0.00 $_{ 0.00}$ &  0.00 $_{ 0.00}$\\
Mult &  34.41 $_{ 0.17}$ &  65.57 $_{ 0.17}$ &  0.01 $_{ 0.01}$ &  0.01 $_{ 0.01}$\\
Conv &  32.98 $_{ 0.27}$ &  66.14 $_{ 0.11}$ &  0.54 $_{ 0.24}$ &  0.34 $_{ 0.10}$\\
TL &  49.06 $_{ 0.55}$ &  49.44 $_{ 0.33}$ &  1.07 $_{ 0.64}$ &  0.44 $_{ 0.27}$\\
RF &  53.09 $_{ 0.46}$ &  46.18 $_{ 0.32}$ &  0.48 $_{ 0.14}$ &  0.26 $_{ 0.08}$\\

\bottomrule

    \end{tabular}
    }
    \caption{Percentages assigned to each type of error for the relational task. 
     We report the average error proportions for all the methods on 5 random seeds and the standard error.}
    \label{tab:rel_err}
\end{table}

Table \ref{tab:rel_err} gives a breakdown of errors. Recall that the distractors have a specific structure: if a correct caption for the image is \lang{a$\mathcal{R}$b}, then the given distractors are: \lang{b$\mathcal{R}$a}, \lang{a$\mathcal{S}$b}, \lang{a$\mathcal{R}$c}, \lang{c$\mathcal{R}$b}. 
We note that CLIP, CSP, and TL have a very similar pattern of errors: each model is able to distinguish objects perfectly, and almost all errors are split between \lang{b$\mathcal{R}$a} and \lang{a$\mathcal{S}$b} - tuples that have been seen in training.
The three commutative models, Add, Mult, and Conv, also have a distinctive error pattern. Errors are again focused on \lang{b$\mathcal{R}$a} and \lang{a$\mathcal{S}$b}, with approximately a 1:2 split. This indicates that the models select the relation $\mathcal{R}$ 50\% of the time, and $\mathcal{S}$ the other 50\%. When $\mathcal{R}$ is selected, the predictions are split again between \lang{a$\mathcal{R}$b} and \lang{b$\mathcal{R}$a}, since these cannot be distinguished by the commutative models.
Although the overall performance of RF is similar to these models, the pattern of errors is more similar to that of  CLIP, CSP, and TL.
Finally, CLIP-FT has another different pattern of errors, in which more of the error is now on the objects, rather than the relation. 
We also note that these errors are much noisier than for the CDSMs.

 \section{Discussion}
Our work highlights the limitations of CLIP as a basis for compositional language representations. 
We show that CLIP is capable of disassociating objects and adjectives, enabling it to behave compositionally in the single-object setting. However, it appears to lack a richer structure necessary for compositions that require more abstraction, such as syntax-sensitive variable binding. 
We find that fine-tuning CLIP or training composition-aware models (CDSMs) does not help the model generalize better on the unseen classes for two-object and relation settings.
Our results show that among the CLIP variants, CLIP-FT overfits to the training set and achieves high training accuracy while hurting the generalization accuracy. 
CSP can show improved training accuracy over CLIP and sometimes show increases in validation and generalization accuracy but not always.
Among the syntax insensitive models, we see that Add, Mult, and Conv improve on the training accuracy on the single-object and the two-object settings but only Add generalizes to held-out classes in the single-object setting.
As expected, these models cannot represent order and achieve accuracy close to chance on the relational dataset. 
Our results with type-logical models (TL) have high training accuracy but validation and generalization accuracy are usually close to 0. 
Finally, RF can learn to generalize to classes in the single-object dataset but achieves chance on the two-object and the relational dataset. 
Our experiments focus only on CLIP, and thus should be interpreted conservatively. 
Newer visual encoders trained with different training objectives may produce better results, even with the same text encoders we use in the paper. 
Or, perhaps, progress on compositionality both in visual and text encoding will be necessary to alleviate the problems highlighted here. 
Overall, our results motivate the need for pretraining methods in VLMs that account for binding for better compositionality. 

We also shed light on the benchmarking datasets used in compositional zero-shot learning. 
Typical benchmarking datasets for this task are MIT-States~\citep{isola:cvpr15}, UT-Zappos~\citep{yu:cvpr14}, and C-GQA~\citep{mancini:cvpr21}.
CLIP and CSP show strong performance compared to several existing methods on these datasets (see Section 5 in \citet{nayak2022}).
However, these datasets do not explicitly test for binding of adjectives to nouns, i.e., they are restricted to a single-object setting. 
While this setting captures one important aspect of composition, it does not require models to encode an abstract, order-aware syntax, a critical component of linguistic composition.
In our experiments, we find that CLIP and CSP show high accuracy on the single-object dataset (Section \ref{sec:single_obj}) but the performance drops dramatically on the two-object dataset (Section \ref{sec:two_object}) and relational dataset (Section \ref{sec:rel}).
Challenging datasets like ARO \citep{yuksekgonul:iclr23} show that fine-tuning CLIP with harder negative images and captions can improve CLIP's accuracy on the relational split that accounts for the order of objects.
Our training setup shares similarities as we include hard negative captions for each image.
However, we do not see improved performance after fine-tuning.
Recent work~\citep{hsieh:neurips23} shows that the ARO benchmark includes test examples that can be solved without the visual encoder which could explain the possible improvement in performance. 
These findings motivate the need for more realistic and challenging benchmarks that test for binding and order.

\section{Related Work}
\paragraph{Compositionality in Language}
Our work contributes to the extensive body of work in compositionality and language spanning several decades~\citep{smolensky1990,plate1995,baroni2010,coecke2010,socher2012,mccoy2019,smolensky:ai2022}.
Key models of composition used in language include simple elementwise composition \citep{mitchell2010}, neural models of composition \citep{socher2012}, type-logical models of composition \citep{baroni2010,coecke2010}, and role-filler modes of composition \citep{smolensky1990,plate1995,mccoy2019}. 
We focus on type-logical and role-filler models of composition.
In the area of type-logical models, our work extends models from \citet{maillard2015,wijnholds2020,nagarajan2018} to learn from both images and text and to handle a wider range of compositions.
Within the area of role-filler approaches, recent work has looked at approaches to reasoning \citep{chen2020}, mathematics \citep{russin2021}, and whether recurrent neural networks can be emulated using role-filler approaches \citep{mccoy2019}. 
In particular, \citet{mccoy2019} use tensor product representations to show that sentence encoders~\citep{conneau2017a,kiros2015} can be well approximated by a ``bag of words'' model. 
In this work, we show that CLIP image embeddings behave like a  ``bag of concepts''.

\paragraph{Compositionality in Vision}
There is a growing interest in compositionality and vision  \citep{misra:cvpr17,nagarajan2018,naeem:cvpr21,mancini:cvpr21,lovering2022unit,nayak2022,yun:arxiv22,tull2023}.
Several architectures have been proposed to improve benchmark results on compositional zero-shot learning datasets~\citep{yu:cvpr14,isola:cvpr15,mancini:cvpr21}. 
However, these datasets are often restricted to an adjective-noun setting, ignoring concept binding. 
Recently, datasets such as CREPE~\citep{ma:arxiv22}, ARO~\citep{yuksekgonul:iclr23}, and Winoground~\citep{thrush:cvpr22} study compositionality in VLMs including concept binding, but may not provide a faithful and controlled environment benchmark \citep{hsieh:neurips23}.
In contrast, we build a controlled setup without potential confounders that arise with real-world images to carefully study compositional visual reasoning. 
Concurrently, \citet{clark:arxiv23} compared the performance of frozen CLIP and Imagen, a text-to-image model, on a task similar to our two-object dataset. 
They find that Imagen, in some cases, performs more strongly, suggesting that generative models are better at binding concepts.

\section{Conclusion}
We investigate the ability of CLIP and variants and CDSMs in a controlled environment to perform compositional visual reasoning tasks.
Our results show that CLIP performs well on the single adjective-noun compositions but struggles on compositional tasks that rely on the ability to bind variables. 
Some of the CDSMs perform well on single adjective-noun composition but show performance closer to chance in the two-object and relational tasks. 
Our work not only sheds light on the limitations of CLIP but also suggests that the pretraining of VLMs should account for binding and order for better compositional generalization. 

\section{Limitations and Risk}
\subsection{Models}
We run our experiments on one major VLM (CLIP) and compare these results with a set of compositional models. 
Results on the benchmarking datasets we propose may differ for other VLMs. 
The compositional models we test do not include some types of model such as Recursive Neural Networks \citep{socher2012}, but we do compare key types of model (type-logical and role-filler) from the compositional literature.

\subsection{Datasets}
The Concept Binding Benchmark that we propose studies concept binding with artificially generated shapes. 
While the simplicity of our datasets strengthens the findings, we suspect that the results may differ with more realistic images.

\subsection{Language}
The language we look at is limited to English. 
For the CLIP models that we use, we are limited to English, however, for the compositional models, it would be possible to use other languages, including alternative grammatical structures and word orderings. 
The kind of language used in the labels is very simple, and further work could include more complicated descriptions of the images.

\subsection{Risk}
This research presents limited risk, due to the abstract nature of the datasets and the limited domain of investigation. All previously existing artefacts have been used within the limits of their original purpose.

\section{Ethical Considerations}
The abstract nature of the datasets we use means that ethical implications of the type of modeling done are minimal. We do use English as a language, however, the methods we propose for the CDSMs could be applied to other languages, as in \citet{moortgat2017lexical}. The training methodology involves fine-tuning a VLM with a large number of parameters (see Table \ref{tab:trainable}), however use of this model can be minimized by saving out frozen image embeddings and using these to train CDSMs.

\section*{Acknowledgements}
We thank Beth Pearson for sharing helpful code snippets to run the BLIP experiments.  
ML carried out this work during a visit to the LUNAR group at Brown, and thanks EP and members of the group for invaluable discussion and input. 
NN, PY, and SB make the following acknowledgements. 
This material is based on research sponsored by Defense Advanced Research Projects Agency (DARPA)
and Air Force Research Laboratory (AFRL) under
agreement number FA8750-19-2-1006. The U.S. Government is authorized to reproduce and distribute
reprints for Governmental purposes notwithstanding
any copyright notation thereon. The views and conclusions contained herein are those of the authors and
should not be interpreted as necessarily representing
the official policies or endorsements, either expressed or implied, of Defense Advanced Research Projects Agency (DARPA) and Air Force Research Laboratory
(AFRL) or the U.S. Government. We gratefully acknowledge support from Google and Cisco. Disclosure:
Stephen Bach is an advisor to Snorkel AI, a company that provides software and services for weakly supervised machine learning.

\bibliography{compositional_images}
\bibliographystyle{acl_natbib}

\clearpage
\appendix

\section{Training Details}
\label{sec:training_details}
We provide the training details and hyperparameters used in the experiments. 
We build the training and evaluation pipeline in PyTorch \citep{paszke:neurips19}.
The models are trained on a single NVIDIA RTX 3090, A40, or V100 GPU depending on their availability.
The models are trained for 20 epochs which takes about 6-20 minutes per epoch depending on the dataset. 
Table \ref{tab:trainable} shows the number of trainable parameters in all the models used in our experiment. 

We have three categories of models: CLIP, CLIP variants, and CDSMs (Add, Mult, Conv, TL, RF). 
All the models use pre-trained CLIP ViT-L/14 in the experiments \footnote{\href{https://github.com/openai/CLIP/blob/main/model-card.md}{https://github.com/openai/CLIP/blob/main/model-card.md}.}. 
These methods except CLIP are trained with a cross entropy loss on the train split using an Adam optimizer. 
We use frozen CLIP to predict the classes for the images in the datasets.
During training, we set the batch size of $32$ and weight decay of $10^{-5}$.
CLIP (FT) fine-tunes all the model parameters including the vision and text encoder with a learning rate of $10^{-7}$.
In CSP, we initialize the token embeddings by averaging the embeddings of all the tokens in the English name of the adjective, noun, or relation to get one initial token embedding per concept.
Then, we fine-tune them on the training split with a learning rate of $10^{-6}$.
In CDSMs, we randomly initialize the model parameters and train them with a learning rate of $5 \cdot 10^{-4}$.
We train all our models on the train split and use the validation split to select the final model for testing based on accuracy.

\begin{table}[h]
    \centering
    \begin{tabular}{lcc}\toprule
     &  \multicolumn{2}{c}{Dataset}\\\cmidrule{2-3}
     Method & Single/Two-object & Relational\\\midrule
    CLIP-FT & 429M & 429M\\
    CSP & 8,448 & 5,376 \\
    Add & 8,448 & 5,376\\
    Mult & 8,448 & 5,376\\
    Conv & 8,448 & 5,376 \\
    RF & 9,984 & 7,680 \\
    TL & 4.7M & 2.3M \\\bottomrule
    \end{tabular}
    \caption{The number of trainable parameters in each experiment.}
    \label{tab:trainable}
\end{table}

\section{Training Algorithm}
We describe the algorithm used to train the models.
Models are trained to align the caption vectors with the image vectors.
Algorithm \ref{alg:unified_training} shows the training algorithm for adjective-noun phrases.
We follow a similar procedure to train relational phrases. 

\begin{algorithm}[htbp]
\small
\SetAlgoLined
\SetKwInOut{Input}{Input}\SetKwInOut{Output}{Output}
\Input{Training dataset $\sS$, image encoder $\mathcal{I}$, composition encoder $\mathcal{T}$,  learnable parameters $\vtheta$, adjectives $\sA$, nouns $\sN$, $\lambda$ weight decay, number of distractors $D$, number of epochs $M$}
\Output{The model parameters $\vtheta$}

\For{i $\gets$ 1 to $M$}{
\ForEach{$x,y=(a,n) \in \sS$}{
$\vx \gets \mathcal{I}(x)$; get the image vector\\
$\sY_{\mathrm{neg}}^{D} \gets$ sample $D$ distractors from $\sY_{neg}=\sY\setminus \{y\}$\\
$l_{\mathrm{pos}}\gets\vx\cdot\mathcal{T}(a,n)$\\
$l_{\mathrm{neg}}\gets \sum_{y_{\mathrm{neg}}\in\sY_{\mathrm{neg}}^{D}}\vx\cdot\mathcal{T}(y_{\mathrm{neg}})$\\

$p_{\vtheta}(y=(a,n)|x) \gets \frac{\exp{(l_{\mathrm{pos}}})}{\exp{(l_{\mathrm{pos}} + l_{\mathrm{neg}}})}$\\
$\mathcal{L} \gets -\log p_{\vtheta}(y|x) + \lambda ||\vtheta||_{2}$; cross entropy loss with weight decay\\
$\vtheta \gets $ update all learnable parameters
}
}
return $\vtheta$; the learned model parameters
 \caption{Algorithm to train the model on the adjective-noun compositions.}
 \label{alg:unified_training}
\end{algorithm}

\section{Calibrated Stacking}\label{sec:calibrated_stacking}
Calibrated stacking is a standard practice in zero-shot learning~\citep{chao:eccv16,nayak:tmlr22}.
Zero-shot models tend to be overconfident or biased towards seen classes because they only see the unseen classes as negatives or they are excluded from the training altogether. 
We can fix this overconfidence by simply calibrating the predictions on validation data. 
Following prior work in zero-shot learning, we add a calibration coefficient to lower the cosine similarity score of the seen classes.
During testing, we use the calibration coefficient and calculate the accuracy. 
\begin{table*}[t]
    \centering
    \begin{tabular}{llrrrlrrrlrrr}\toprule
    &&\multicolumn{3}{c}{Single Object} && \multicolumn{3}{c}{Two Object} && \multicolumn{3}{c}{Relational} \\\cmidrule{3-5}\cmidrule{7-9}\cmidrule{11-13}
     Model && Train & Val. & Gen. && Train & Val. & Gen. && Train & Val. & Gen.  \\\cmidrule{1-1}\cmidrule{3-5}\cmidrule{7-9}\cmidrule{11-13}
       BLIP-Base  && 94.23 & 91.36 & 87.82 && 27.79 & 8.37 & 27.96 && 17.54 & 50.07 & 0.0 \\
       BLIP-Large  && 98.46 & 98.62 & 97.46 && 22.66 & 15.75 & 40.61 && 22.35 & 22.18 & 40.34 \\\bottomrule
    \end{tabular}
    \caption{Results for BLIP on the single-object, two-object, and the relational datasets from the concept binding benchmark. }
    \label{tab:blip}
\end{table*}

\paragraph{Setup}
To test whether calibrated stacking improves generalization accuracy, we experiment with CSP on the single object dataset but modify the train set. 
To find a calibration coefficient, we need a validation set to include seen and unseen classes. 
Since our validation set contains only unseen classes as the positive labels, we need a additional validation set with seen classes. 
To fix this issue, we randomly sample 10\% of the train set and use that as the seen validation set. 
We train our model on the remaining 90\% of the data with the same training details (see Section \ref{sec:experiments}). 
Next, we compute the cosine similarity scores for the seen and the unseen validation sets and search for the calibration coefficient. 
Next, we get the highest cosine similarity $l_{\mathrm{max}}$ and vary the calibration $-l_{\mathrm{max}}$ to $+l_{\mathrm{max}}$ with a step size of $l_{\mathrm{max}}/100$ and choose the coefficient with the highest harmonic mean of the seen and the unseen accuracy.
Finally, we use the calibration coefficient on the generalization set and report the performance. 

\paragraph{Results}
\begin{table}[h]
    \centering
    \begin{tabular}{lr}\toprule
    Method & Generalization \\\midrule
    CLIP & 92.39\\
    CSP & 88.74\\
    CSP + calib. & 96.31\\\bottomrule
    \end{tabular}
    \caption{The results for single-object setting on the generalization split. 
    For CSP and CSP + calib., we report the average accuracy on 5 random seeds.}
    \label{tab:calibration}
\end{table}
Table \ref{tab:calibration} shows that CSP with calibration improves by 8 points on the generalization split. 
We also see that CSP improves over CLIP by 4 points showing that the model has learned to generalize to unseen adjective-noun compositions. 
This shows that fine-tuned models, including the CSDMs, could potentially generalize better than frozen CLIP with calibration. 
These results are in line with the literature in compositional zero-shot learning that calibrate the predictions and show improved results on the adjective-noun datasets~\citep{purushwalkam:iccv19,ruis:neurips21}. 
However, we find that calibrating the predictions in the two-object setting does not improve the generalization performance the same way. 
This may be due to the construction of the two-object dataset. 
In the validation split we have the classes \lang{brown cube} and \lang{green sphere}. 
The ``hard distractors'' for these classes are \lang{brown sphere} and \lang{green cube}. However, these hard distractors come from the generalization set, i.e., they are unseen classes. 
This means the calibration does not decrease the cosine similarity of the hard distractors, making it difficult to calibrate the validation set.
Finally, calibration is not applicable to the relational dataset because we consider only two classes in the generalization split, \lang{cube behind cylinder} and \lang{cylinder behind cube}, that are equivalent. 
This means, we only see one class at a time and simply setting the probability of the distractors to 0, we can get 100\% accuracy on the generalization set. 
For this reason, we do not calibrate on the relational dataset and leave the experiment for the future. 

\section{Experiments with BLIP}
We further highlight the limitations of contrastive vision-language models by evaluating BLIP~\citep{li2022blip} on the concept binding benchmark. 
BLIP is a pretrained vision-language model trained with a unimodal image encoder, unimodal text encoder, image-grounded text encoder, and image-grounded text decoder.
We consider two BLIP model sizes: BLIP-Base and BLIP-Large. 
We follow the same evaluation procedure used for CLIP. 

Table \ref{tab:blip} shows the results for BLIP on the concept binding benchmark. 
Our results are similar to CLIP across all the datasets. 
On the single object datasets, we find that BLIP achieves good performance on all the splits. 
However, we find the performance of both the models dramatically reduces on the two-object and relational datasets. 
This further highlights the grounded compositionality problem in vision-language models.

\section{License}
All the code including the models and the datasets used in this work are released under open-source licenses. 
Blender is released under the GNU GPL License, CLIP is released under the MIT license, and CSP is released under the BSD-3 license. 
We have released the code and concept binding benchmark dataset under the Apache 2 license. 

\end{document}